\documentclass[conference]{IEEEtran}
\IEEEoverridecommandlockouts
\usepackage{cite}
\usepackage{amsmath,amssymb,amsfonts}
\usepackage{algorithmic}
\usepackage{graphicx}
\usepackage{textcomp}
\def\BibTeX{{\rm B\kern-.05em{\sc i\kern-.025em b}\kern-.08em
    T\kern-.1667em\lower.7ex\hbox{E}\kern-.125emX}}

\usepackage{tikz}

\newcommand{\dd}[2]{\frac{\partial #1}{\partial #2}}
\newcommand{\xa}{x_a}
\newcommand{\xp}{x_p}
\newcommand{\cat}{\theta}
\newcommand{\Ra}{r_a}
\newcommand{\Rp}{r_p}

\newcommand{\eqname}{}

\usepackage{hyperref}
\newcommand\copyrighttext{%
  \footnotesize \textcopyright 2021 IEEE. Personal use of this material is permitted. Permission from IEEE must be obtained for all other uses, in any current or future media, including reprinting/republishing this material for advertising or promotional purposes, creating new collective works, for resale or redistribution to servers or lists, or reuse of any copyrighted component of this work in other works. 2021 IEEE 19th International Conference on Industrial Informatics (INDIN),
  DOI: \href{https://ieeexplore.ieee.org/document/9557382}{10.1109/INDIN45523.2021.9557382}}
\newcommand\copyrightnotice{%
\begin{tikzpicture}[remember picture,overlay]
\node[anchor=south,yshift=10pt] at (current page.south) {\fbox{\parbox{\dimexpr\textwidth-\fboxsep-\fboxrule\relax}{\copyrighttext}}};
\end{tikzpicture}%
}

\begin{document}

\title{A Grid-Structured Model of Tubular Reactors
}

\author{\IEEEauthorblockN{Katsiaryna Haitsiukevich}
\IEEEauthorblockA{
\textit{Aalto University}\\
Espoo, Finland \\
\{firstname.lastname\}@aalto.fi}
\and
\IEEEauthorblockN{Samuli Bergman, Cesar de Araujo Filho}
\IEEEauthorblockA{
\textit{Neste}, 
Espoo, Finland \\
\{firstname.lastname\}@neste.com \\
\{cesar.araujo\}@neste.com}
\and
\IEEEauthorblockN{Francesco Corona, Alexander Ilin}
\IEEEauthorblockA{
\textit{Aalto University}\\
Espoo, Finland \\
\{firstname.lastname\}@aalto.fi}
}

\maketitle
\copyrightnotice

\begin{abstract}
We propose a grid-like computational model of tubular reactors. The architecture is inspired by the computations performed by solvers of partial differential equations which describe the dynamics of the chemical process inside a tubular reactor. The proposed model may be entirely based on the known form of the partial differential equations or it may contain generic machine learning components such as multi-layer perceptrons. We show that the proposed model can be trained using limited amounts of data to describe the state of a fixed-bed catalytic reactor. The trained model can reconstruct unmeasured states such as the catalyst activity using the measurements of inlet concentrations and temperatures along the reactor.
\end{abstract}


\section{Introduction}
Tubular reactors are commonly used in process industry for producing value-added products by operating a chemical reaction. During operation, the reactor is constantly monitored to control the process inside the reactor, to detect malfunctioning or to schedule maintenance.
It is typical that many variables describing the state of the reactor
cannot be directly measured  during operation and they must be estimated using measured variables. This is a classical state estimation problem which requires a dynamic model of the reactor.

There is extensive literature on modeling tubular reactors using first principles, e.g. \cite{froment2010chemical}, \cite{davis2012fundamentals}. The chemical process inside the reactor is usually described by partial differential equations (PDEs). The parameters of the PDEs are determined by conservation laws and in the ideal world they would fit the observed data well. Unfortunately, due to inaccuracies in modeling assumptions and high level of noise, the first-principles models often do not agree with the data.
This motivates building reactor models in a data-driven way by fitting generic parametric models such as neural networks 
\cite{goodfellow2016deep} to process observations.
This is a viable solution in case one can collect large amounts of observations, otherwise the problem of overfitting becomes inevitable. Unfortunately, data scarcity is a common problem in this domain.

In this work, we assume that the tubular reactor system can be described by PDEs whose form is known but which contain some unknown nonlinearities.
We propose a model whose computational graph is based on the knowledge of how PDEs are discretized and solved. The proposed model is a deep model with a grid structure in which each cell performs computations inspired by the known form of the PDEs. Each cell may optionally contain generic neural networks to model unknown nonlinear components that can be present in the PDEs. The model parameters are estimated using gradient descent and backpropagation, which is common for deep learning models.
We use simulated data to show that our model can learn to recover the state of a chemical process in a fixed bed reactor with catalyst poisoning. We apply the same model to data collected from a real chemical reactor and show that the model provides a good fit.

\section{Fixed bed reactor with catalyst poisoning}

As an example tubular reactor, we consider a fixed bed reactor with exothermic reaction
$
    \text{A} + \text{H} \rightarrow \text{B},
$
in which a set of aromatic compounds collectively denoted by A are hydrogenated on the surface of the catalyst to produce a set of de-aromatized compounds B \cite{price1977catalyst}.
We assume that the catalyst deactivates irreversibly due to adsorption of poisoning agent P, meaning that the catalyst activity is decreasing over time. We also assume that there is a constant abundance of H.

A macroscopic characterization of this type of reactors in space and time $t$ is given by a set of PDEs accounting for mass and energy balances. We restrict ourselves to a single spatial dimension, the axis of the reactor $z$. We also assume that the reactor operates under plug flow conditions, that is the diffusion phenomena are negligible, and consider only convection and reaction phenomena. 

The concentration $\xa(z,t)$ of aromatic compounds A in the reactor is given as a solution to the mass balance equation
\begin{align} 
\dd{\xa}{t} &= -U \dd{\xa}{z} - \alpha(T) \Ra,
\label{eq:xa}
\end{align}
where $U$ is the constant velocity of the fluid along the reactor and $\Ra$ is the reaction rate at which A is converted into B. The term $\alpha(T)$ denotes a linear function of temperature $T$. 

Similarly, the concentration $\xp(z,t)$ of poison P solves
\begin{align} 
\dd{\xp}{t} &= -U \dd{\xp}{z} - \alpha(T) \Rp,
\label{eq:xp}
\end{align}
where $\Rp$ is the rate at which P is adsorbed on the catalyst.

Under the assumption that the reactor is adiabatic, values $T(z,t)$ of the temperature function solve the energy balance
\begin{align} 
\dd{T}{t} &= - \beta(\xa, T) U \dd{T}{z} + \gamma \Ra,
\label{eq:T}
\end{align}
where $\beta(\xa, T)$ is proportional to $\xa/T$ and $\gamma$ is a constant.

To account for decay in catalyst activity, $\cat(z,t)$ solves
\begin{align} 
\frac{d \theta}{dt} & = -r_d
\,,
\label{eq:cat}
\end{align}
where  $r_d$ is the deactivation rate.
For all locations, the catalyst activity starts at one %
and decays to zero with time. 

The rate $r_a$ of the hydrogenation reaction is
\begin{equation}
  \Ra = \frac{k_0 K_0 \exp\left(\frac{-Q -E}{R T}\right)}
         {1 + K_0 \exp\left(- \frac{Q}{RT}\right) P \xa^{l_2}}
  P^2 \xa^{l_1} x_h \cat
\,,
\label{eq:Ra}
\end{equation}
where $Q$ and $E$ are adsorption and hydrogenation activation energies, $k_0$ is pre-exponential factor for hydrogenation, $K_0$ is aromatics adsorption constant, $l_1$ and $l_2$ are exponents constants, $P$ is pressure, $R$ is gas constant and the concentration of hydrogen $x_h$ which is assumed to be constant.
The rate $r_p$ at the which the poisoning agent P is adsorbed onto the catalyst and the catalyst deactivation rate $r_d$ are related:
\begin{align}
r_d &= k^0_d \exp\left(-E_d/(R T)\right) P \xp \cat
\\
\Rp &= r_d C_d \cat,
\label{eq:Rp}
\end{align}
where $E_d$ is activation energy for the catalyst poisoning, $k^0_d$ is poisoning pre-exponential factor and $C_d$ is the catalyst poison adsorption capacity.

\begin{figure}[t]
\centering
\includegraphics[width=\linewidth,trim={5mm 6mm 3mm 3mm},clip]{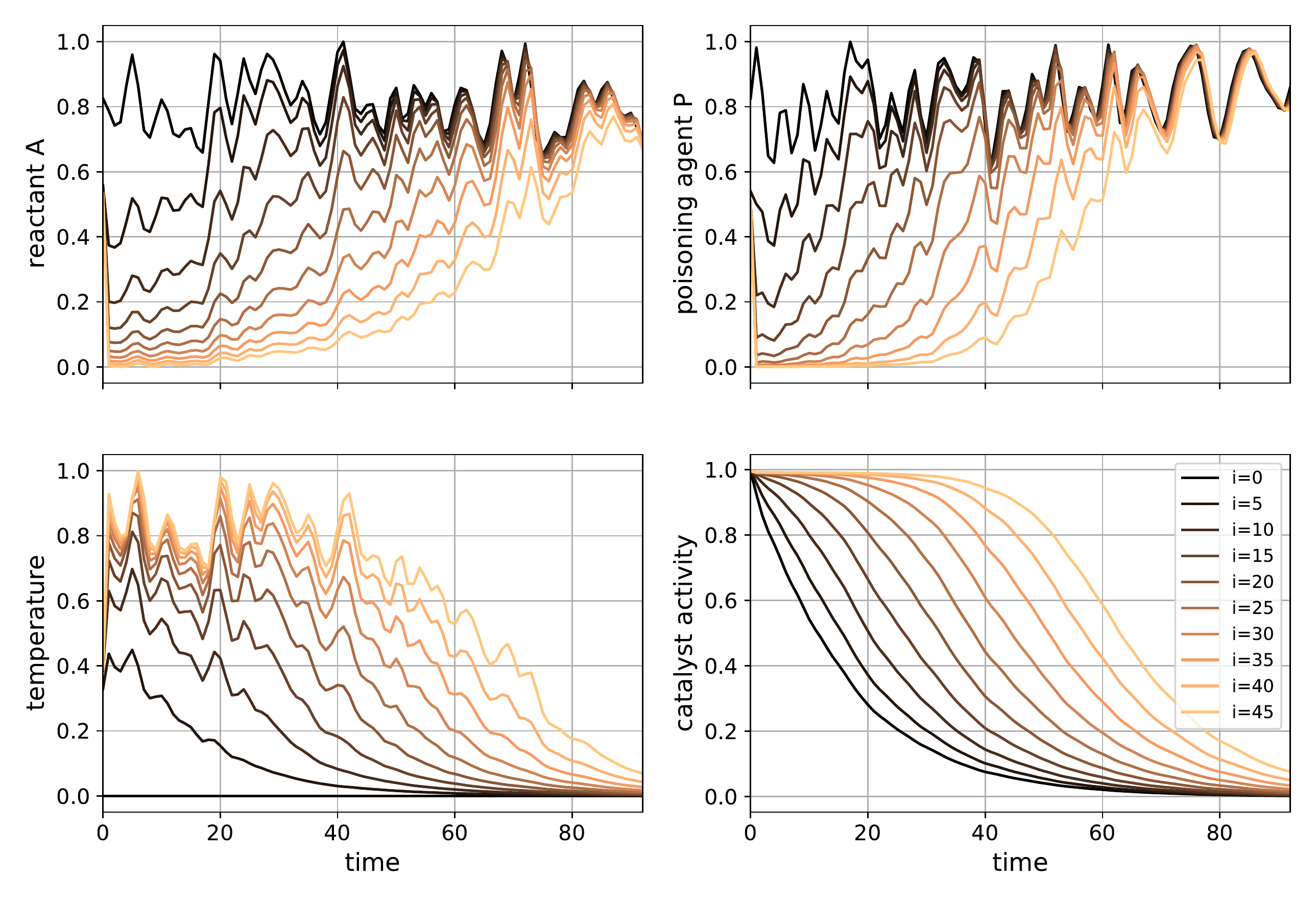}
\caption{One catalyst cycle of the simulated data.
Each color represents one reactor level $i$. The signals are scaled to the range $[0, 1]$.
}
\label{f:gen_data}
\end{figure}

For a given set of initial and boundary conditions, \figurename~\ref{f:gen_data} shows the time and space evolution of the system variables over one catalyst cycle. At the beginning of the cycle, the reaction and catalyst deactivation occur close to the reactor inlet, as indicated by the large spatial gradients of $\xa$, $T$ and $\xp$. Later, catalyst poisoning causes the shift of the reaction towards the reactor outlet. When the catalyst is deactivated, the concentrations $\xa$, $\xp$ and the temperatures $T$ do not change any longer and the reactor behaves like a tubular pipe.

\section{Grid-structured model for tubular reactors}

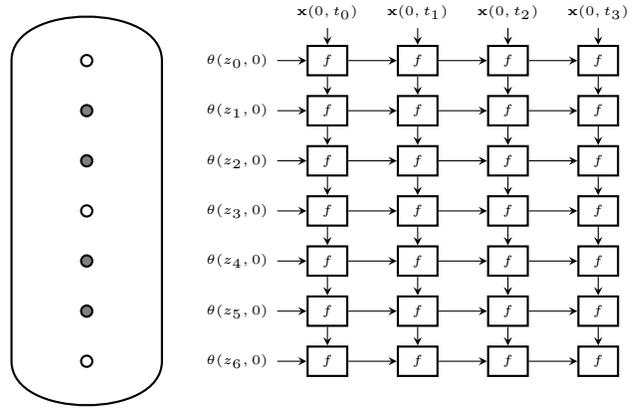
\begin{figure}[t]
\centering
\tikzset{%
  block/.style   = {draw, thick, rectangle, minimum height=1.em, minimum width=1.5em},
  obs/.style     = {draw, thick, shape=circle, inner sep=1.5pt},
  dscr/.style    = {draw, thick, shape=circle, inner sep=1.5pt, fill=gray}
}

\begin{tikzpicture}[auto, node distance=5mm]

\draw
[thick] (1,0) to [out=90,in=90] (-1, 0)
[thick] (1.,0) -- (1.,-4)
[thick] (-1.,0) -- (-1.,-4)
[thick] (1,-4) to [out=270,in=270] (-1, -4)
;

\foreach \z in {0,2,...,4}{
  \draw node at (0, -\z) [obs] {};
}
\foreach \z in {0.666,1.3333,2.666,3.3333}{
  \draw node at (0, -\z) [dscr] {};
}

\foreach \z in {0,...,6}
{
  \foreach \t in {0,...,3}
  {
    \draw node at (3.2+\t*1.2, -\z*2/3) [block] (b\t\z){\tiny $f$};
  }
}

\foreach \t in {0,...,3}
{
  \draw node at (3.2+\t*1.2, 0.65) (in\t) {\tiny $\mathbf{x}(0, t_\t)$};
  \draw [->,>=stealth] (in\t) -- (b\t0);

  \foreach \z [count=\zn] in {0,...,5}
  {
    \draw [->,>=stealth] (b\t\z) -- (b\t\zn);
  }
}

\foreach \z in {0,...,6}
{
  \draw node at (2, -\z*2/3) (cat\z) {\tiny $\cat(z_\z, 0)$};
  \draw [->,>=stealth] (cat\z) -- (b0\z);
  \foreach \tp [count=\t] in {0,...,2}
  {
    \draw [->,>=stealth] (b\tp\z) -- (b\t\z);
  }
}
\end{tikzpicture}
\caption{Left: A plan of a tubular reactor with the locations of the temperature sensors. Black circles indicate the location without sensors. Right: The computational
graph of the proposed model where $\mathbf{x}$ denotes a vector with components $\xa, \xp, T$.}
\label{f:model}
\end{figure}

To design a surrogate neural model of the reactor, we assume that the process develops slowly in time, that is the reactor is in a pseudo-steady-state. We set the time derivatives $\dd{\xa}{t}$, $\dd{\xp}{t}$, $\dd{T}{t}$ to zero which modifies \eqref{eq:xa}--\eqref{eq:cat} to a set of ordinary differential equations (ODEs).
Assuming that we know the concentrations $\xa(z_0, t)$, $\xp(z_0, t)$ and temperatures $T(z_0, t)$ at the reactor inlet and that initial catalyst activity $\cat(z, t_0) = 1$, we can predict the concentrations, temperatures and the catalyst activity along the reactor by solving the resulting set of ODEs using the Euler's method:
\begin{align} 
x(z + \Delta z,t) &\approx x(z,t) + f_x(\xa, \xp, T, \cat) \Delta z
\label{eq:cell_x}
\\
\cat(z, t+\Delta t) &\approx \cat(z,t) - r_d \Delta t
\,,
\label{eq:cell_cat}
\end{align}
where $x(z,t)$ denotes one of the three variables $\xa$, $\xp$ or $T$ and functions $f_x$ are derived from \eqref{eq:xa}--\eqref{eq:T} with 
partial derivatives $\dd{\xa}{t}$, $\dd{\xp}{t}$, $\dd{T}{t}$ set to zero.
These computations can be implemented by a grid-structured computational graph shown in \figurename~\ref{f:model}.
Each cell in that graph corresponds to one point $(z, t)$ in a chosen discretization of space and time. The inputs to the model are the boundary and initial conditions. The vertical top-down connections implement computations in \eqref{eq:cell_x} while the horizontal connections represent \eqref{eq:cell_cat}. The parameters of each cell are defined by the parameters of the original PDEs \eqref{eq:xa}--\eqref{eq:cat}.
The parameters are shared for all cells to guarantee the required property of equivariance to translations in space and time.

Suppose now that the model is unknown and we need to fit it to process observations. We assume that we can measure the concentrations $\xa(z_0, t)$, $\xp(z_0, t)$ at the inlet and the temperatures $T^*(z_i, t_j)$ at locations with indices $i \in I_z$ along the reactor at times with indices $j \in I_t$. We can tune the model by building a computational graph in \figurename~\ref{f:model} and by fitting the model outputs
to the temperature observations. This can be done by minimizing the mean-squared error loss
\begin{align}
  C = \frac{1}{|I_z| |I_t|} \sum_{i \in I_z} \sum_{t \in I_t} (T(z_i, t_j) - T^*(z_i, t_j))^2
\,.
\label{eq:loss}
\end{align}

The proposed architecture can be seen as a deep recurrent model in which each cell corresponds one location in space and time. Tuning the parameters of the model can be challenging due to vanishing and exploding gradients, like in other deep learning models. Standard recurrent units such as gated recurrent unit (GRU) \cite{cho2014learning} or long-short term memory (LSTM) \cite{hochreiter1997long}, which are designed to mitigate the vanishing/exploding gradients problem, can be used as computational units in the proposed model. However, this comes with the price of losing the model interpretability and requiring much richer data to cover all possible operating conditions of the reactor.

\section{Related work}

Several neural networks have architectures resembling or motivated by numerical solvers of differential equations (see, e.g., \cite{gonzalez1998identification,lu2018beyond,long2018pde,belbute2020combining}). Many works built data-driven models by combining generic neural networks with PDE and ODE solvers (see, e.g., \cite{belbute2020combining,chen2018neuralode}).
Our model can be viewed as a particular instance of interactions networks \cite{battaglia2018in} in which each level of the reactor interacts with the neighboring levels. Instead of using generic function approximators to model the interactions, we limit the family of functions to achieve interpretability and sample efficiency.

In the chemical engineering literature, several works proposed grid-structured architectures for modeling reactors \cite{deans1960computational,guo2008cellmodel}.
The model proposed in \cite{deans1960computational} was used for predicting the mixing characteristics of fixed beds of spheres
in fixed-bed catalystic reactors.
A similar approach was used in \cite{guo2008cellmodel} to model the behavior of trickle-bed reactors employed for the exothermic hydrotreating of benzene. They described the steady-state of the reactor using one-dimensional or two-dimensional networks in which dimensions corresponded to the geometrical dimensions of the reactor.
Three model parameters in \cite{guo2008cellmodel} were learned from the data while the rest of the parameters were taken from literature. In our approach, the dimensions of the grid correspond to space and time. We learn all the model parameters in a gradient-based optimization.

\section{Experiments with simulated data}

\subsection{Synthetic data}
\label{sec:simulations}

We first conducted experiments on synthetic data generated by solving PDEs~\eqref{eq:xa}--\eqref{eq:cat} numerically. To simulate the data, we used constant temperature at the inlet, while the inlet concentrations were varied (see \figurename~\ref{f:gen_data}).
The generated data were downsampled such that one trajectory representing one catalyst cycle contained 93 time points at 46 reactor levels.
\figurename~\ref{f:gen_data} shows the simulated data with one catalyst cycle that were used for training.

The inputs of the model were the concentrations and the temperature at the reactor inlet and temperature measurements $T(z_j, t)$ at locations $0, 4, 8, \dots, 44$ were used as the targets.
Temperatures at locations $1, 5, \dots, 41$ were used as validation set 1. Temperatures at the remaining locations were used as validation set 2 that we used for model selection. Note that the validation sets significantly correlate with the training data because all data come from the same catalyst cycle. We nevertheless used such validation sets because in practice the modeler often has measurements of only one catalyst cycle. We tested the model performance on five catalyst cycles simulated with the following scenarios.
In Test~1, we used the same operating conditions as in the training data but different random seed.
In Test~2, the inlet temperature was increased by 7\% compared to the training scenario.
In Test~3, the inlet concentrations $\xa$ were increased by 20\%.
In Test~4, the inlet concentrations $\xp$ were decreased by 15\%.
Test~5 combines the changes from Tests 2--4. 

\subsection{Training details}\label{sec:synt_res}

We trained the proposed grid-structured model in which each cell
implemented the following computations:
\begin{align} 
\xa^\text{out} &= \xa - \frac{k_a}{U} g_a(T, \xa) \cat
\label{eq:cell_a}
\\ 
\xp^\text{out} &= \xp - \frac{k_p}{U} T g_p(T, \xp) \cat^2
\label{eq:cell_p}
\\
T^\text{out} &= T + \frac{k_T g_a(T, \xa)}{U(k_1 + k_2 \xa)}  \cat
\label{eq:cell_T}
\\
\cat^\text{out} &= \cat - k_\theta g_p(T, \xp) \cat
\,,
\label{eq:cell_th}
\end{align}
where $\xa, \xp, T, \cat$ are the cell inputs,
$\xa^\text{out}, \xp^\text{out}, T^\text{out}, \cat^\text{out}$
are the cell outputs and $k_a, k_p, k_T, k_\cat, k_1, k_2$ are the model parameters.
The functional form of the cell was derived from the PDEs \eqref{eq:xa}--\eqref{eq:cat}.
The important property of this cell is that the same functions
$g_a$, $g_p$ are used in \eqref{eq:cell_a}, \eqref{eq:cell_T} and \eqref{eq:cell_p}, \eqref{eq:cell_th}, respectively. This couples the dynamics of the four variables and makes it possible to reconstruct the unobserved variables using the observed ones.

\begin{table*}[htp] 
\caption{Normalized RMSE of temperature reconstructions in experiments with simulated data.}
\label{tab:mse}
\begin{center}\begin{tabular} {l | c  c  c | c  c  c | c  c  c  c  c }
\hline
& \multicolumn{3}{c|}{$T$, training cycle} & \multicolumn{3}{c|}{$\Delta T$, training cycle} & \multicolumn{5}{c}{$\Delta T$, test cycles}
\\
\cline{2-12}
& Train & Valid.\ set 1 & Valid.\ set 2 & Train & Valid.\ set 1 & Valid.\ set 2 & Test 1 & Test 2 & Test 3 & Test 4 & Test 5
\\
\hline
GRU with 8 neurons & \textbf{0.04} & -- & -- & \textbf{0.13} & -- & -- & 0.57 & 1.34 & 1.04 & 2.26 & 2.12
\\
Grid-GRU & 0.10 & 2.62 & \textbf{0.12} & 0.19 & 5.86 & 0.67 & 0.28 & 0.83 & 0.78 & 0.74 & 1.58 
\\
Grid-GRU with augm & 0.14 & 0.14 & 0.14 & 0.25 & 0.35 & 0.30 & 0.27 & 0.99 & 0.36 & 0.43 & 1.26
\\
PDE param\ & 0.15 & 0.14 & 0.14 & 0.23 & 0.23 & 0.24 & 0.25 & \textbf{0.35} & 0.27 & \textbf{0.26} & 0.39
\\
MLP & 0.14 & 0.13 & 0.13 & 0.16 & \textbf{0.15} & \textbf{0.16} & \textbf{0.19} & 0.54 & 0.36 & 0.33 & 0.54
\\
MLP-reg & 0.13 & \textbf{0.12} & \textbf{0.12} & 0.17 & 0.17 & 0.17 & 0.21 & 0.38 & \textbf{0.26} & 0.27 & \textbf{0.38}
\\
\hline
\end{tabular} \end{center}
\end{table*}

In the first version of the cell, the functional forms of $g_a$ and $g_p$ was dictated by the original PDE model $g_a(T, \xa) = e^{-k_3/T} T \xa$, $g_p(T, \xp) = e^{- k_4 / T} \xp$.
We refer to this version of the cell as \textit{PDE param}. 
In the second version of the cell, we assumed that the exact forms of the reaction and adsorption rates were not known and therefore functions $g_a$, $g_p$ were modeled with generic MLP networks.
The MLPs had the architectures 32-64-32-1, 32-32-1 for $g_a$ and $g_p$ respectively, with layer normalization \cite{ba2016layernorm} followed by ReLU non-linearity after each hidden layer and softplus nonlinearity in the output layer. The MLP inputs were normalized with min-max normalization.
We denote this cell \textit{MLP} in the tables with the results. 

When training the model with the MLP-based cell, we optionally used two extra penalty terms in the loss function:
\begin{align*}
C_1 &= \lambda_{a} \sum_{i,j} \min(0, \xa(z_i, t_j))^2 + \lambda_{p} \sum_{i,j} \min(0, \xp(z_i, t_j))^2
\\
C_2 &= \lambda_{a} \sum_{t_j < t'} \xa(z_N, t_j)^2 + \lambda_{p} \sum_{t_j < t'} \xp(z_N, t_j)^2
\,.
\end{align*}
The first penalty term forces the non-negativity of the concentrations
while the second term forces the outlet concentrations at the beginning of the catalyst cycle to be close to zero. We used hyperparameters $\lambda_{a}=1$ and $\lambda_{p}=100$ and $t'=46$. We denote the model trained with such regularization as \textit{MLP-reg}.%

All MLP-based models were trained with the Adam optimizer \cite{kingma2017adam} for 1000 gradient updates with learning rate $10^{-3}$ and then tuned for 200 more updates with learning rate $10^{-4}$. The model with the PDE parametrization was trained with Adam with the same schedule but with ten times larger rates.

As the first baseline model, we used a generic GRU with eight neurons. The inputs of the GRU model were the inlet concentrations, the inlet temperature and velocity $U$. Naturally, this model can predict temperatures only at the same locations which were used in the training set.
The second baseline was a grid model with the same computational graph as in \figurename~\ref{f:model}, in which each cell was implemented by a generic GRU with eight neurons. Compared to the simple GRU baseline, this model has an in-built property of spatial equivariance. 
The inputs of both GRU-based models were normalized. %
Both baseline models were trained for 3000 epochs with learning rate $10^{-3}$ and then for 3000 epochs more with learning rate $10^{-4}$.

\subsection{Results}

Table~\ref{tab:mse} shows the root mean-squared errors normalized by the data standard deviation (NRMSE) for the temperature reconstructions obtained with the trained models. We present the quality of the reconstructions of temperatures T and temperature differences between two subsequent sensor locations in a given set:
\begin{align}
    \Delta T(z_i, t) = T(z_{i}, t) - T(z_{j_i}, t)
\,,
\label{eq:deltaT}
\end{align}
where $z_i$ and $z_{j_i}$ are the locations of successive sensors.
For example, for the training set we compute temperature differences between locations 4 and 0, 8 and 4 and so on. $\Delta T$ is an indicative metric because it reflects the reaction rate at a particular location. The best model for each class was selected from five runs with different seeds based on the MSE of
$\Delta T$ reconstruction on validation set 2.

The GRU baseline model severely overfits to the training data and  completely fails to generalize to the test cycles due to the lack of the right inductive bias.
The grid-GRU model trained in a standard way finds a solution that overfits to the regular spacing of the temperature sensors. This is indicated by the large error on validation set 1. The networks learns to produce meaningful reconstructions only at the locations of the measurements, thus using intermediate locations of the grid as hidden layers that can develop arbitrary representations.

To mitigate this problem, we trained the grid-GRU model with the following data augmentation scheme. At each training iteration, we randomized the locations selected from the model prediction vector for the loss calculation by \eqname\eqref{eq:loss}. Each location was shifted independently either to the previous level (with probability 0.25) or to the next level (with probability 0.25) or kept at the correct location. The model trained in this way is denoted as \textit{Grid-GRU with augm}. The results show that the Grid-GRU model trained with augmentations was able to generalize much better. The model performed well on Tests 1, 3 and 4 but failed on Tests 2 and 5, that is the model could not generalize to temperature changes.

The proposed grid-structured models showed very good performance on all of the five test sets. The best performance was obtained with the PDE-based cell. This model used the correct analytical form of the reaction and poisoning rates, which let it extrapolate well outside the training distribution.
The model with a generic MLP cell performed very well too and it was able to outperform the PDE-based model on some tests.

\begin{table}[htp]
\caption{Average $\bar{R}^2$ score for $\xa$, $\xp$, $T$ and $\cat$ reconstructions}
\label{tab:r2}
\centering
\begin{tabular}{@{}lcccccc@{}}
\hline
Model & Train & Test 1 & Test 2 & Test 3 & Test 4 & Test 5
\\
\hline
PDE param & 0.963 & \textbf{0.958} & \textbf{0.959} & 0.959 & \textbf{0.957} & \textbf{0.956}
\\
MLP & 0.953 & 0.943 & 0.945 & 0.955 & 0.895 & 0.913
\\
MLP-reg & \textbf{0.967} & 0.957 & 0.954 & \textbf{0.960} & 0.942 & 0.945
\\
\hline
\end{tabular}
\end{table}

\begin{figure}[t]
\centering
\includegraphics[width=\linewidth,trim={10mm 17mm 5mm 2mm},clip]{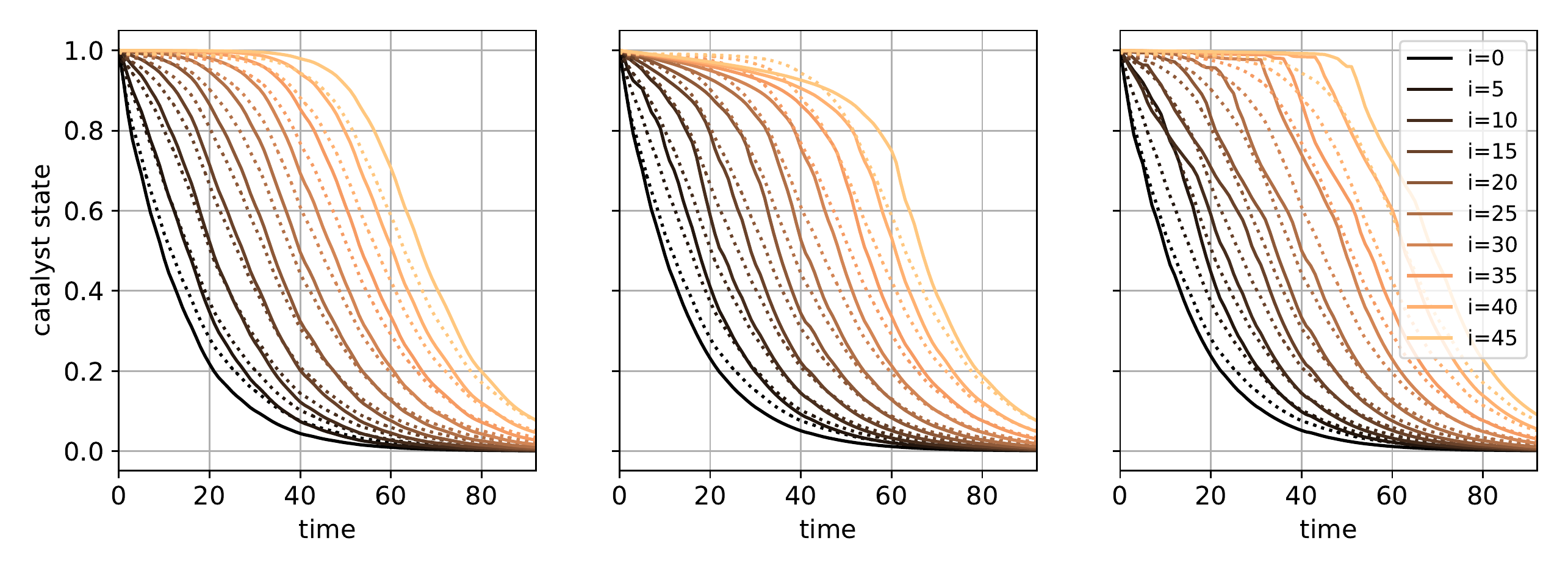}
\\
\includegraphics[width=\linewidth,trim={5mm 5mm 5mm 2mm},clip]{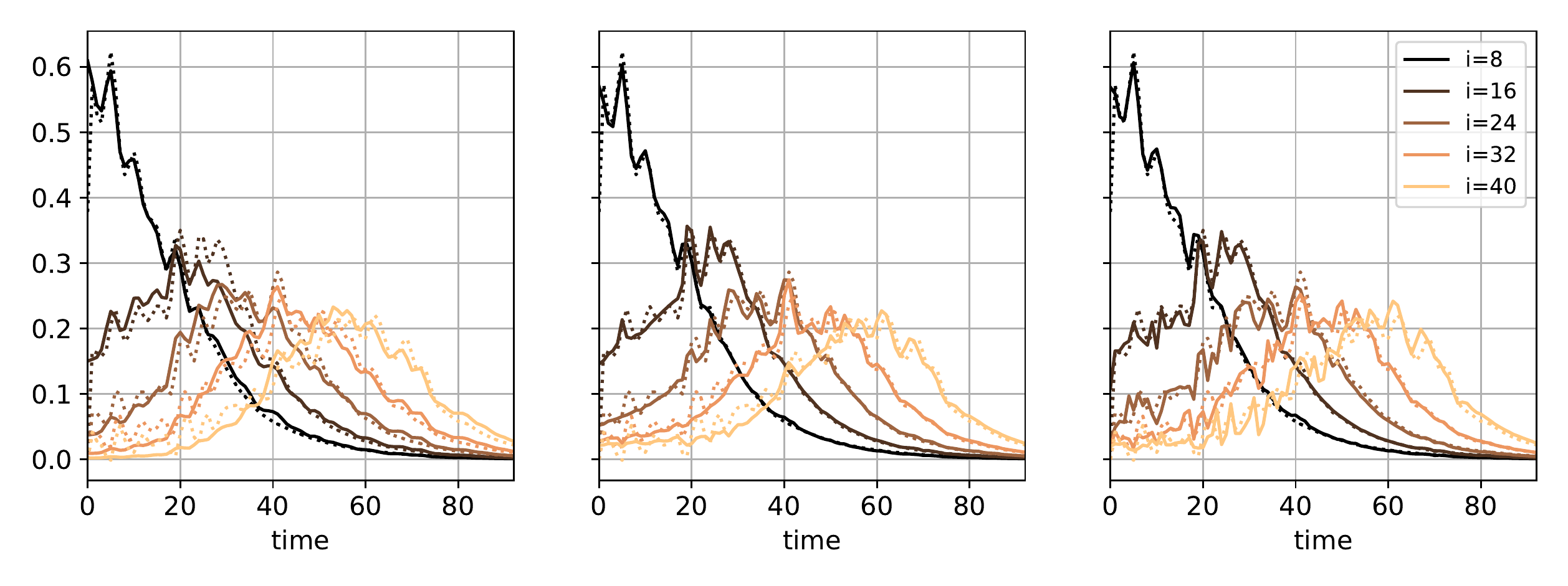}
\caption{Experiments with simulated data: The catalyst activity (above) and the temperature differences $\Delta T$ (below) reconstructed at different reactor levels by the proposed grid-structured models. Left: Cell with PDE parameterization, middle: MLP-based cell, right: MLP-based cell with regularization. The dotted lines represent the simulated ground truth. The $\Delta T$ time series are min-max scaled.}
\label{f:cat_profiles}
\end{figure}

In Table~\ref{tab:r2}, we compare the accuracy of the grid-structured models in the task of predicting the values of the unobserved variables in all spatial and temporal locations. As the metric, we use the average of $R^2$ scores for four system states. %
Note that the table contains results only for the proposed grid models because the baseline models cannot reconstruct unobserved system states. The high values of the $R^2$-score indicate that both models produced excellent reconstructions. The PDE-based model is generally slightly better than the MLP-based model.
\figurename~\ref{f:cat_profiles} shows the reconstructions of the catalyst activity and
the temperature differences $\Delta T$, as defined in \eqname\eqref{eq:deltaT},
at different reactor levels. One can see that the produced estimates are close to the ground truth.

\section{Experiments with real data}

We applied the proposed models to estimate the states of a large-scale hydrotreatment unit. The unit is used for continuous saturation of hydrocarbons using hydrogen.
The reductive hydrogenation occurs on the surface of a single bed of a catalyst. The reaction releases heat and the process is assumed to be adiabatic.
The presence of a poisoning component and other impurities leads to irreversible catalyst deactivation. Thus, the catalyst needs to be periodically replaced. %
The considered unit is equipped with two sequential reactors of the same nature, the outlet of the first reactor is directly fed to the second one. The unit is monitored by taking the following measurements:
\begin{itemize}
\item the flow rate and the temperature of the hydrocarbon reactor feed, measured in real time;

\item the feed composition including the concentration of reactants and the poisoning agent, measured periodically by laboratory analysis;

\item the temperatures inside the reactors, measured in real time by an array of thermocouples (sensors) installed inside the catalyst bed.

\end{itemize}

We used data from a period which represented one life cycle of the catalyst.
Since our focus was on modeling slowly changing phenomena such as catalyst deactivation, we took daily averages of temperature measurements and downsampled them to create time series $T(z_i, t)$ with the same sampling frequency as the laboratory measurements of the concentrations $\xa(z_0, t)$ and $\xp(z_0, t)$. Linear velocity $U(t)$ was approximated by the flow rate of the hydrocarbon feed.

We built a model of the whole unit which included both reactors. We discretized the dimension along the flow direction such that the discretization points were close to the locations of the temperature sensors. This resulted in 46 locations $z_i$,
where $i=0$ and $i=21$  were the locations of the inlets of the first and the second reactors, respectively.
The thermocouples were installed at locations $i \in \{8, 16, 20\}$ in the first reactor and at the following locations of the second reactor: $i \in \{25, 28, 31, 34, 37, 40, 43, 45\}$ along pole~1, $i \in \{24, 27, 30, 33, 36, 39, 42, 45\}$ along pole~2 and $i\in \{23, 26, 29, 32, 35, 38, 41, 45\}$ along pole~3.

We assume that the considered hydrotreatment unit is described well by \eqref{eq:xa}--\eqref{eq:cat} and we modeled its pre-processed data with the proposed grid-structured models.
The inputs of the models were inlet concentrations $\xa(z_0, t)$, $\xp(z_0, t)$, inlet temperatures $T(z_0, t)$ and linear velocity $U(t)$. The models were trained to fit temperature measurements $T(z_i, t)$. As the training set, we used the measurements from the first reactor and the sensors attached to the first pole of the second reactor. We used two validation sets which consisted of the temperature measurements from the second and the third poles of the second reactor.

We trained four grid-structured models on the reactor data. The first model was Grid-GRU with 32 neurons trained with learning rates $10^{-3}$ and $10^{-4}$ for 15000 iterations each.
We did not use the data augmentation that we used for training the same model on the simulated data. This regularization was not needed because the irregular placement of the temperature sensors prevented the type of overfitting that we observed for the simulated data.
The other models that we trained were based on the cell defined in \eqname\eqref{eq:cell_a}--\eqref{eq:cell_th}. These models were trained with the same hyperparameters as in the experiments with the simulated data.
The only difference was that we used data augmentation by applying additive Gaussian noise to all measurements. The noise had zero mean and the standard deviation was 0.002 times the mean of the corrupted variable.

\begin{figure}[t]
\hspace{0.3mm}
\includegraphics[width=.988\linewidth,trim={10mm 17mm 5mm 2mm},clip]{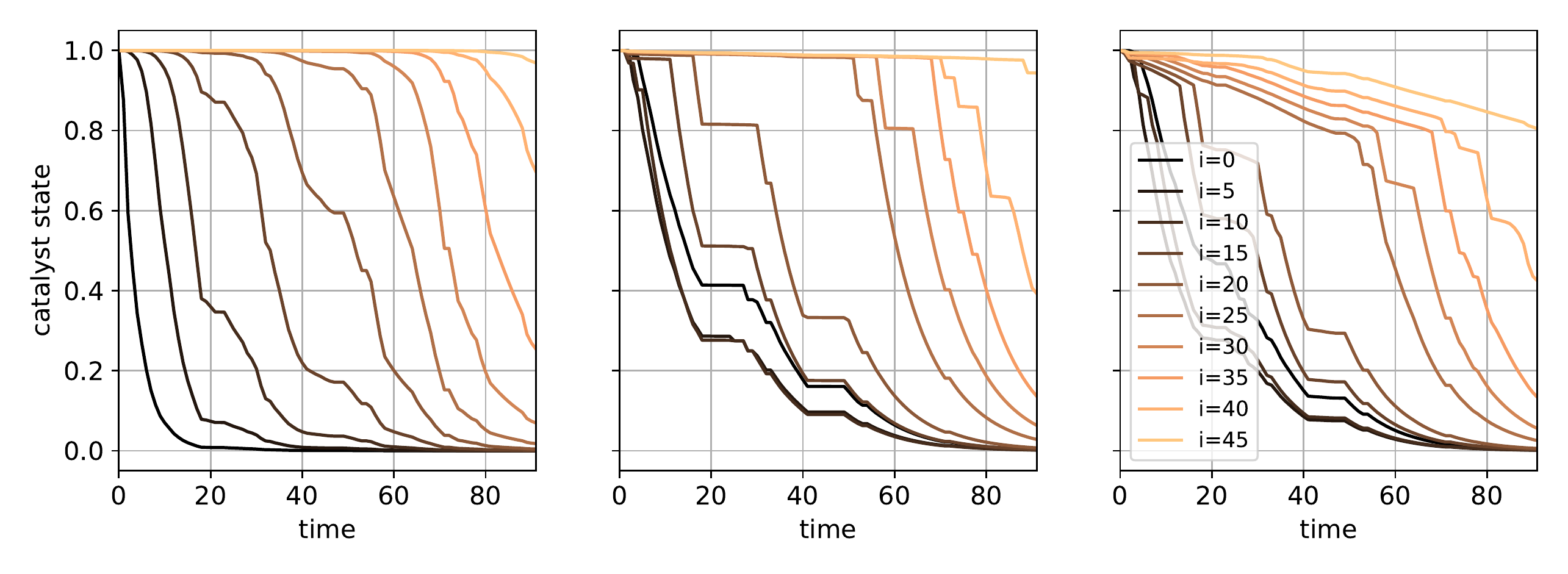}
\\
\includegraphics[width=\linewidth,trim={4mm 5mm 5mm 2mm},clip]{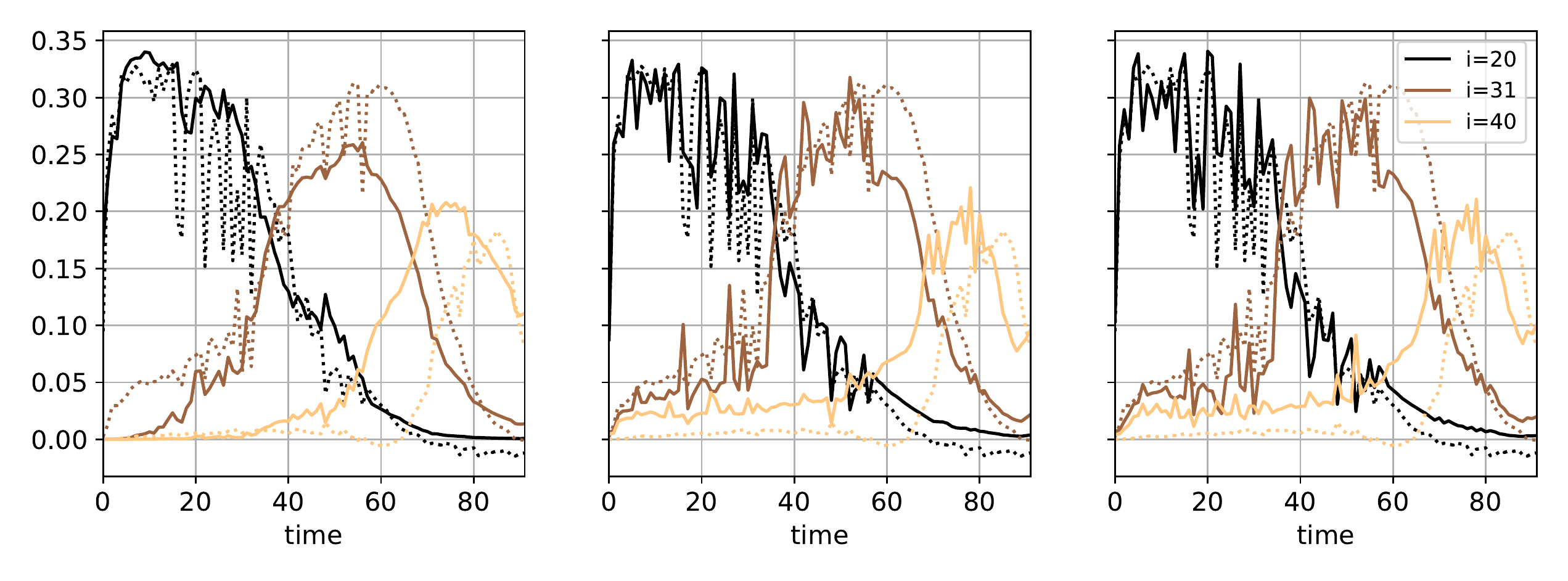}
\caption{Experiments with real data: The catalyst activity (above) and the temperature differences $\Delta T$ (below) reconstructed at different reactor levels by the proposed grid-structured models. Left: Cell with PDE parameterization, middle: MLP-based cell, right: MLP-based cell with regularization. The dotted lines represent the measurements. The $\Delta T$ time series are min-max scaled.}
\label{f:dt_real}
\end{figure}

\begin{table}[t]
\caption{Pearson correlation coefficient of temperature reconstructions
for a real reactor.}
\centering
\begin{tabular}[tbp]{@{}l|ccc@{}}
\hline
& Train & Valid.~set~1 & Valid.~set~2 \\
\hline
Grid-GRU & 0.909 & 0.845 & 0.860 \\
PDE param & 0.797 & 0.757 & 0.805 \\
MLP & 0.834 & 0.791 & 0.798 \\
MLP-reg  & 0.821 & 0.785 & 0.800 \\
\hline
\end{tabular}
\label{tab:model_real}
\end{table}

\figurename~\ref{f:dt_real} present the estimated catalyst activity and the fit of the temperature differences $\Delta T$, as defined in \eqname\eqref{eq:deltaT}, which were obtained by the proposed models. One can see that both models capture the drift of the reaction from the inlet towards the outlet of the reactor. The MLP-based model provides a reasonable fit to the data and is comparable with the results obtained with the PDE parametrization.
Table~\ref{tab:model_real} presents the Pearson correlation coefficient of the temperature predictions
on the training and two validation sets. These results suggest that all the models can generalize to the validation sets from the same catalyst cycle reasonably well.
We did not test the accuracy of the models on a data set with different inputs because %
of data scarcity.

\begin{figure}[t]
\centering
\includegraphics[width=\linewidth,trim={5mm 5mm 5mm 2mm},clip]{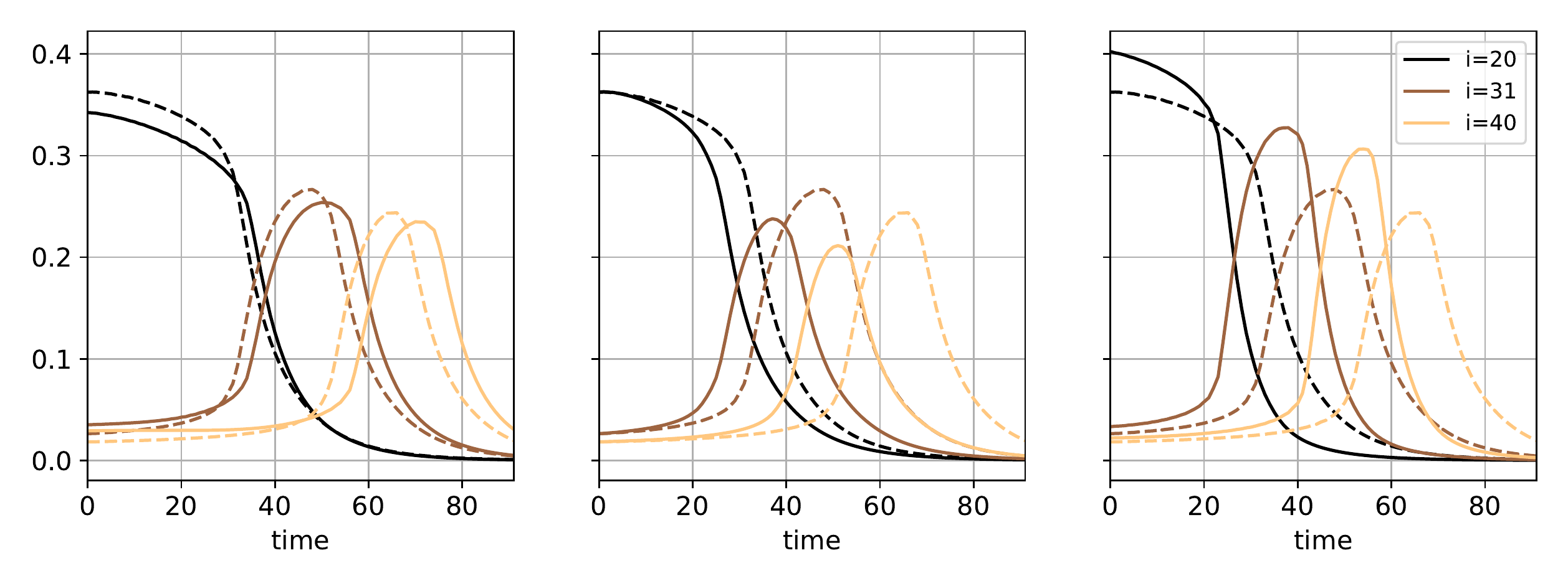}
\caption{Responses of the trained grid model with the MLP-based cell to changes of the operating conditions. 
Dashed lines represent temperature differences $\Delta T$ predicted by the model for reference operating conditions. The solid lines represent model predictions under changed operating conditions: increase of the inlet temperature by 10\% (left), increase of the inlet concentration of P by 25\% (middle), increase of the inlet concentration of A by 25\% (right).}
\label{f:dt_const_sensi_t}
\end{figure}

In \figurename~\ref{f:dt_const_sensi_t}, we demonstrate how the models respond to changes of the inputs compared to reference operating conditions. We can see that the increase of the inlet temperature slows down deactivation while the increase of the inlet concentrations of P and A speeds up deactivation, as in \eqref{eq:xa}--\eqref{eq:cat}.

\section{Conclusion}

In this paper, we proposed a model that can be used to estimate unobserved states of tubular reactors. The model has a grid-structured architecture which is motivated by the computations performed when solving partial differential equations which describe chemical processes inside such reactors. The model can be viewed as a recurrent network whose computational units are constructed using the prior knowledge about the process dynamics. %
We demonstrate that the trained model can be used to monitor variables, such as catalyst activity, which can be costly or impossible to measure directly in real-world industrial reactors.

One direction of future research is to use the proposed architecture for monitoring process states using new observations. This can be done, for example, by training the grid-structured model with a generic computational unit such as GRU using data produced by process simulators under randomized scenarios \cite{tobin2017domain}. 

Due to data scarcity in many industrial domains, there is clear need for more research on combining first-principles models with the data-driven approach. In this paper, we introduced the right inductive bias into a data-driven model by structuring its computational graph, selecting the right computations and by using certain data augmentation schemes. Other promising approaches include learning a combined model containing both a physical component and a data-driven component \cite{guen2020augmenting} or by learning data-driven models that respect conservation laws such as mass or energy balance \cite{greydanus2019hamiltonian, cranmer2020lagrangian}.

\section*{Acknowledgment}
We would like to thank Akshaya Athwale, Muhammad Emzir, Sakira Hassan, Simo S\"arkk\"a, James Kabugo, Teemu Ikonen, Viljami Iso-Markku, Stefan T\"otterman, Amir Shirdel, Sanna Laitinen and Kristian Bergman for help in the data gathering process and fruitful discussions. We thank CSC (IT Center for Science, Finland) for computational resources and the Academy of Finland for the support within the Flagship programme: Finnish Center for Artificial Intelligence (FCAI).

\bibliographystyle{IEEEtran}
\bibliography{IEEEabrv,references}

\end{document}